# MobileDenseAttn: A Dual-Stream Architecture for Accurate and Interpretable Brain Tumor Detection

Shudipta Banik
Department of CSE, East Delta University, Bangladesh
Email: shudiptobanik99@gmail.com

Muna Das
Department of CSE, USTC, Bangladesh
Email: munadas76145@gmail.com

Trapa Banik
Department of EEE, University of Southern Indiana, USA
Email: tbanik@usi.edu

Md. Ehsanul Haque
Department of CSE, East West University, Bangladesh
Email: ehsanulhaquesohan758@gmail.com

*Abstract*—The detection of brain tumor in MRI is an important aspect of ensuring timely diagnostics and treatment; however, manual analysis is commonly long and error-prone. Current approaches are not universal because they have limited generalization to heterogeneous tumors, are computationally inefficient, are not interpretable, and lack transparency, thus limiting trustworthiness. To overcome these issues, we introduce *MobileDenseAttn*, a fusion model of dual streams of MobileNetV2 and DenseNet201 that can help gradually improve the feature representation scale, computing efficiency, and visual explanations via Grad-CAM. Our model uses feature-level fusion and is trained on an augmented dataset of 6,020 MRI scans representing glioma, meningioma, pituitary tumors, and normal samples. Measured under strict 5-fold cross-validation protocols, *MobileDenseAttn* provides a training accuracy of 99.75%, a testing accuracy of 98.35%, and a stable F1-score of 0.9835 (95% CI: 0.9743–0.9920). The extensive validation shows the stability of the model, and the comparative analysis proves that it is a great advancement over the baseline models (VGG19, DenseNet201, MobileNetV2) with a +3.67% accuracy increase and a 39.3% decrease in training time compared to VGG19. The Grad-CAM heatmaps clearly show tumor-affected areas, offering clinically significant localization and improving interpretability. These findings position *MobileDenseAttn* as an efficient, high-performance, interpretable model with a high probability of becoming a clinically practical tool in identifying brain tumors in the real world.

*Index Terms*—Brain tumor detection, Ensemble Learning, PMRAM dataset, Brain tumor classification, Explainable AI

## I. INTRODUCTION

The diagnosis of brain tumors is one of the most complex and important tasks of medical diagnosis. The growth of these tumors is usually silent, and they show few or no symptoms until they are in their advanced phases. The rising rates of brain cancer around the world, of which the World Health Organization reported more than 300,000 every year, demonstrates that early detection is necessary [1], [2]. The survival chances in case of early recognition are outrageous, yet there is a major problem of detecting the tumors in their initial stages without any symptoms. Conventional methods of manual detection are slow, inaccurate, and ineffective, especially in cases where large amounts of MRI scan information are to be analyzed, mainly consisting of manual detection by expert radiologists [3]–[6].

Regardless of the medical imaging progress, the current automated techniques have multiple limitations, such as accuracy, generalization, computational efficiency, and interpretability. A large number of existing methods lack the ability to generalize to new tumor types or imaging conditions, resulting in suboptimal performance in practical applications. The need to create more stable, explainable, and computationally powerful brain tumor detection mechanisms, therefore, exists. Although recent studies have come up with some gains on this area, research gaps still exist. The majority of models are based on single-streams architectures, thus becoming under-generalized and lacking the means to capture the complex features. The absence of explainable AI (XAI) approaches also limits transparency and it is not easy to trust predictions in clinical situations [7].The model architectures are generally not innovative, as most solutions operate within established frameworks without exploring dual-stream or hybrid models that may better capture complex patterns in multifaceted data [8].. Additionally, one can see that cross-validation performance can be ignored, and models are usually not stable when an alternate sub-sample of the data is used. Lastly, the computational cost is not adequately addressed, and most models are inefficient and are not applicable in large clinical set up. In addition, multi-stage pipelines and collection algorithms are also multi-sided and it can be hard to generalize in real-world clinical practice [9].

### A. How We Overcome These Challenges

To solve these problems, we will introduce a hybrid deep leaning design that merges strong points of several neural network architecture. The core of our methodology is to enhance generalization, increase computational efficiency and offer interpretability. Regarding generalization, we apply feature fusion methods to make sure that the system could be easily adjusted to various tumor types. This increases its usefulness in many clinical settings. When it comes to computational efficiency, our architecture balances the concept of accuracy with speed, which allows it to be used with real-time applications even in such settings where computational resources are scarce. We lastly add interpretability mechanisms, like Grad-CAM, that assist in highlighting and displaying how the model came to its decisions. This is transparent and helps the clinicians know how the model detects zones of tumor in the MRI scans so as to instill value in its forecasts.

The key contributions of this research include:
- The new approach is a combination of effective computation and highly-efficient feature extraction in the detection of brain tumors through dual-stream fusion model.
- The system generalizes well on different tumors and imaging conditions and can be expected to do so well in the real-life clinical context.
- Offers a visualisation of how all the variables might interact in the model making clinicians confident and open.
- Extensive evaluation on a large variety of data set proving to be more accurate in some most important measures accuracy, precision, recall and F1-score.
- The architecture also performs well computationally, albeit it is rather accurate, and thus can be deployed on low resource devices.
- The 5-fold cross-validation will validate the performance of the model and provide reasonable reliability and consistency to it.

Such contributions hold great promise towards the design of more stable, effective and explainable brain tumor detection systems, whose steps towards clinical applicability and real-world implementation are closer.

## Literature Review

Recent studies have been making strides towards detection of brain tumors through deep learning methodologies on MRI and CT scans.

Aamir et al. [10] introduced a hyperparameter-based CNN architecture that generated an average overall accuracy of 97% across three publicly available datasets in an article that also presented the enhanced feature extraction but pointed to challenges such as susceptibility to overfitting and data demands. Likewise, Patel et al. [11] proposed a CNN that combined machine learning (ML) and deep learning (DL), with an accuracy of 92.86%, and balanced precision and recall, prioritizing quality limitations on the used datasets. In another study, Iftikhar et al. [12] increased the accuracy and interpretability of classification by using explainable contexts MLP and CNN with Grad-CAM, SHAP, and LIME translating to 95% test accuracy and recommended expanded dataset and XAI technique investigation. Singh et al. [13] proposed a CNN named BrainNet using transfer learning with networks such as VGG variants and InceptionResV2 and achieved higher testing accuracy of 97.71 % than that of benchmarks, but warned that the results were not generalizable. Involutional Neural Networks (InvNets) with location-adapted kernels have been presented by Asiri et al. [14] and achieve 92% accuracy with a decrease of parameters, compared to traditional CNNs.

Mahmud et al. [15] trained a CNN based on the learning of 3,264 MRI images and demonstrated better results compared to ResNet-50, VGG16, and Inception V3 with 93.3% of validation accuracy and 98.43% AUC, further reaffirming the efficacy of CNNs in early tumor detection. Khaliki et al. [16] tested a variety of models on 2870 MRI images, with VGG16 performing the best with an accuracy of 98%, and transfer learning variants also performing well. Moreover, Wozniac et al. [17] have suggested a Correlation Learning Mechanism (CLM) that uses a combination of CNNs and ANNs on CT scans and segments of the CT scans and achieves an almost 96% accuracy validating the higher level of diagnostic efficiency.

Taken together, such studies shows gradual improvement in terms of accuracy, interpretability, and efficiency of deep learning models to detect brain tumors, but also highlight the importance of heterogeneous data and strong validation to ensure clinical utility.

### B. Gap Identification

Although current literature has shown some improvements, there are various gaps. Single-stream architectures are the basis of most models, restricting the overall versatility of the models and prohibiting the detection of complex features. Also, the absence of explainable AI (XAI) approaches limits transparency, and it is hard to trust a model on predictions in clinical practice. There is also a lack of innovation when it comes to model architecture and the majority of the methods use well-established models and do not consider dual-stream or hybrid architectures whereby they can capture multi-faceted data more adequately. Moreover, there is a common tendency to ignore cross-validation performance and the models do not tend to be stable across the various subsets of data. Lastly, the computational cost has also not been adequately optimized and most models have not been optimized to be efficient and are therefore unreachable to the large scale clinical application. Furthermore, multi-stage pipeline methods and ensemble techniques may be too complex to implement in the real world setting of a clinical setting.

## II. Methodology

This section will provide a thorough account of the methodology that will be used in the research. Figure 1 demonstrates the overall workflow of the proposed approach, showing data acquisition all the way up to the fusion of MobileNetV2 and DenseNet201 into brain tumor detecting, as well as interpretability methods like Grad-CAM etc.

### A. Data Collection and Preprocessing

In this study, the PMRAM Bangladeshi brain cancer dataset was used, containing 1600 raw images in total [18]. In every class, there are 400 raw images which are grouped into the following classes: Pituitary, Meningioma, Glioma, and Normal. A number of preprocessing steps were used in order to improve the features to obtain better classifier performance. To start with, all pictures were normalized having the same size of 224x224 pixels. This was followed by implementation of contrast limited adaptive histogram equalization to provide contrast and increase the details of the images. The Fast Non-Local Means Denoising algorithm was used to optimize the quality of the image and decrease the level of noise. Lastly, an image normalization process was applied to provide consistent input values in the entire dataset, which reduces difficulty of convergence in the training process.

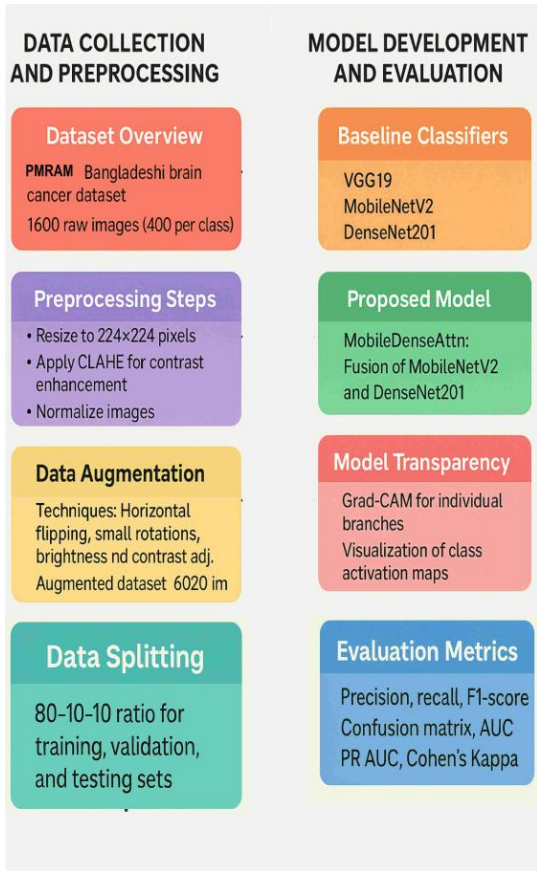

Fig. 1. The proposed workflow for Brain Tumor Detection.

*B. Data Augmentation and Splitting*

To provide more variability to the dataset and strengthen the model, a number of augmentation methods were used. There are 6,020 images in the augmented dataset which were produced by horizontal flipping with a probability of 50%; small-precision rotations of 10 degrees; affine transformation with horizontal flipping and small rotations; small-precision black-white adjustment of brightness and contrast using scaling factor $\alpha = 1.2$ (for contrast) and $\beta = 15$ (for brightness), and (we apply 120%). These methods were selected so that the real-world variability (change in the orientation of an object, light, and noise) could be simulated. Artificially increased dataset allows the model to be more generalized, therefore less likely to overfit and anxious to unseen data.

Upon augmentation, the data set was divided into three parts: training, validation and testing sets in a ratio of 80-10-10. In particular, it was ensured that 80% of the images are added to the training set, 10% to the validation set, and the remaining 10% are added to the testing set. The detailed split info depicted in table I:

TABLE I
DATA SPLIT SUMMARY (TOTAL: 6020 IMAGES)

| Set | Glioma | Meningioma | Pituitary | Normal | Total |
|---|---|---|---|---|---|
| Train | 1193 | 1161 | 1193 | 1267 | 4814 |
| Validation | 149 | 145 | 149 | 158 | 601 |
| Test | 150 | 146 | 150 | 159 | 605 |

*C. Baseline Classifier*

In the domain of image classification, it is essential to turn to the choice of appropriate baseline model to ensure a strong performance. In this paper, we considered three established deep learning architectures, VGG19, MobileNetV2, and DenseNet201, that offer distinct benefits to the table. VGG19 is believed to be one of the deepest and simplest architectures, which by relying on a succession of convolutional layers of small 3x3 filters. Its advantage is that it captures the hierarchical spatial properties hence making it a strong model when detailed representations of images are required. Nonetheless, considering its richness, it comes at a high computational cost to the extent that it necessitates high memory and processing capacity, which in other cases (such as smartphones) may be limited. It uses depthwise separable convolution and inverted residual, which lowers computational demands without lowering accuracy. This model is best suited to real-time applications and has provided a balance between performance and efficiency. In DenseNet201, each layer makes dense connections with every other layer, reusing features in the network. The architecture has a substantial positive impact on gradient flow, and DenseNet201 proves remarkably effective at capturing complex features, in particular where the dataset is limited in size. The capability of taking advantage of all features that are available facilitates its better performance on more complicated classification tasks.

*1) Proposed MobileDenseAttn: A Fusion of MobileNetV2 and DenseNet201:* We introduce MobileDenseAttn, which is the integration between MobileNetV2 and DenseNet201. DenseNet201 and MobileNetV2 are two of the most optimal models currently available in image classification, and by combining them, we target to pick the strong features of both architectures. MobileNetV2 has a depthwise separable convolutions depthwise separable convolutions of feature extractor with a small size. DenseNet201 with its own adequacy of connections propagates the better utilization of features and the improved gradient flow. By combining these models, there is a possibility to utilize all complementary features of the models, leading to an increase in brain tumor detection performance.

*D. Enhance Proposed Model Transparency*

To enhance transparency of the described MobileDenseAttn model, we use Grad-CAM (Gradient-weighted Class Activation Mapping) on each of the branches prior to fusion. Because we cannot directly use Grad-CAM on the fused features we compute the class activation maps separately on MobileNetV2 and DenseNet201. This will aid us to have an idea of the areas affecting the predictions of each model, hence giving a better idea of how both models can aid to the final decision therefore improving the interpretability of the fused architecture.

*E. Evaluation Metrics*

The efficiency of the proposed model is measured by a number of indicators in terms of accuracy, precision, recall, F1-score, Cohen kappa, cross-validation accuracy, computational cost, confidence interval and overall score. The accuracy indicates the correctness of the model as a whole whereas precision will be the number of accurate positive predictions over the total amount of positive predictions. Remember also referred to as sensitivity shows how well the model detects all cases that are relevant. Harmonic mean of precision and recall is balanced against each other and is called the F1-score. Cohen kappa measures the agreement between the values of predicted and actual classifications taking into consideration chance. Cross-validation accuracy gives the mean accuracy result of a k-fold cross-validation. Computational cost can be defined as resources needed to train and perform inference, including time and memory. The statistical reliability of estimates of the metric is identified with confidence intervals. Finally, the total score is used to give a single point of reference to the performance of the models that considers several factors.

### III. RESULTS AND DISCUSSION

As evidenced in Table II, the proposed MobileDenseAttn model was more effective, in addition to being lightweight, on all major metric measures. It demonstrated exceptional learning and generalization abilities since it reached an accuracy of 0.9975 during training, where precision, recall, and F1-score are also equal to 0.9975. MobileDenseAttn was also able to retain the same accuracy of 0.9835 in the testing set, with a precision and recall of 0.9840 and 0.9835 respectively, demonstrating its good balance between false positives and false negatives. Compared to VGG19, which has a lower accuracy score of 0.9468 in the testing set, with precision and recall being strongly unbalanced. Being efficient, MobileNetV2 was not as accurate as MobileDenseAttn with the accuracy on testing being 0.9769 and F1-score being 0.9768. DenseNet201 with training accuracy of 0.9896 and testing accuracy of 0.9636 did not perform well in generalization.

The combination of Attention with MobileNetV2 and DenseNet201 in MobileDenseAttn gives the balance in the efficiency and feature extraction and the final model is the most trustworthy model in brain tumor detection. Also, the 5-fold cross-validation results, depicted at Figure 2, prove the robustness and generalization ability of the model. MobileDenseAttn demonstrates performance and becomes the best option in the realm of clinical experiments in brain tumor classification.

As can be seen in Figure 2 5-fold cross-validation performance on the proposed MobileDenseAttn model is impressive on all measures. The F1-Score shows constant high indicators, reaching the max peek of nearly 0.9894 in fold two, and the overall mean F1-Score of all folds delivers an indicative of 0.9835. The Cohen Kappa also gives a solid agreement alongside the predicted and true labels with a mean value of 0.9784, with very few misclassifications. The accuracy is also always greater than 97% percent, the highest one being 98.96 percent in Fold 2, and the average one is of 0.9838 that reflects on the efficacy and generalizability of the model.

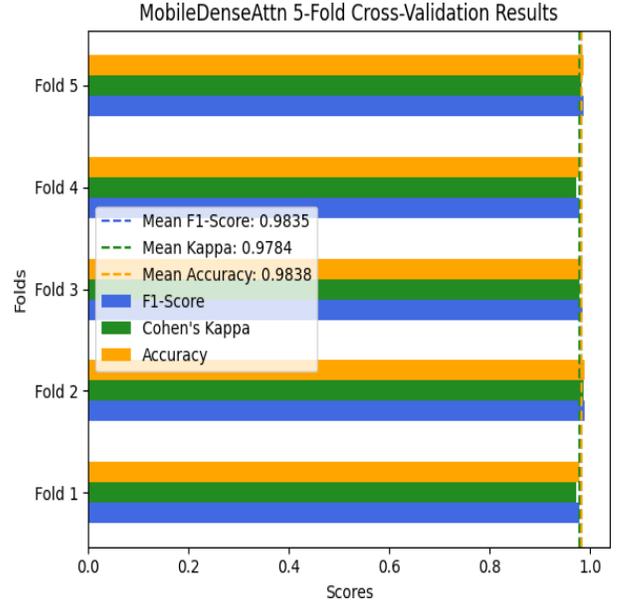

Fig. 2. MobileDenseAttn 5-Fold Cross-Validation Results for Classification Metrics (F1-Score, Cohen's Kappa, Accuracy).

Moreover, in figure 3 The learning curves of MobileDenseAttn, we can see that the model steadily increased its accuracy and loss through all 29 epochs. The accuracy of training rises steadily, peaking around 0.9720 at epoch 23, and validation accuracy rises as well at epoch 24 from 0.9950 to indicate good generalization.

The rate of training loss plunges to the ground, which means that the model is learning well, and the validation loss is also deteriorating with minor spikes here and there, which can be attributed to overfitting towards the end. On the whole, the trend indicates that MobileDenseAttn is effectively learning and generalizing, which makes it suitable when being deployed to use in a real-life scenario.

*A. Computational Efficiency and Inference Performance*

Figure 4 shows computational costs of various models in terms of overall training time and inference time used to train, validate and test the models. Although MobileDenseAttn inherits the advantages of MobileNetV2 and DenseNet201, its overall training time of 913.78 seconds is much shorter than that of VGG19 of 1504.38 seconds and hence more efficient in large-scale training.

TABLE II
PERFORMANCE COMPARISON OF MODELS ON TRAINING AND TESTING SETS

| Model | Training | | | | Testing | | | |
|---|---|---|---|---|---|---|---|---|
| | Accuracy | Precision | Recall | F1-score | Accuracy | Precision | Recall | F1-score |
| VGG19 | 0.9724 | 0.9732 | 0.9724 | 0.9725 | 0.9468 | 0.9481 | 0.9468 | 0.9470 |
| MobileNetV2 | 0.9979 | 0.9979 | 0.9979 | 0.9979 | 0.9769 | 0.9771 | 0.9769 | 0.9768 |
| DenseNet201 | 0.9896 | 0.9897 | 0.9896 | 0.9896 | 0.9636 | 0.9643 | 0.9636 | 0.9636 |
| **MobileDenseAttn** | **0.9975** | **0.9975** | **0.9975** | **0.9975** | **0.9835** | **0.9840** | **0.9835** | **0.9835** |

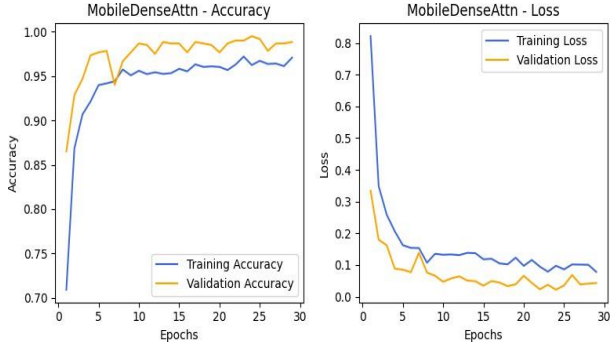

Fig. 3. Validation and Loss Curve of Proposed MobileDenseAttn.

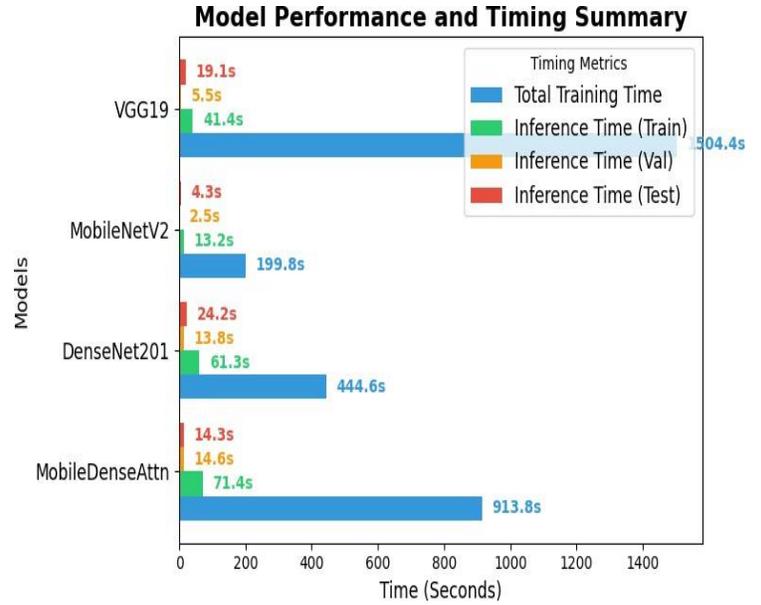

Fig. 4. Computational Cost of Each evaluated model.

In the case of inference, MobileDenseAttn exhibits a competitive performance of 14.31 seconds of testing time, compared with 24.22 seconds of DenseNet201. Naturally, the inference time of the MobileDenseAttn is also better than both VGG19 (19.09 seconds) and MobileNetV2 testing time (4.26 seconds) which is greatly expected by virtue of lightweight design of MobileNetV2.

This combination of high accuracy and low Computational overhead usage makes MobileDenseAttn one of the perfect models to use in a real-time application like brain tumor detection, where accuracy and inference speed are equally important.

### B. Evaluation of the Proposed Model: Confusion Matrix, ROC Curve, and PR AUC

Fig. 5 presents the confusion matrix of the proposed MobileDenseAttn model, which indicates that model performance in terms of brain tumor type differentiation is very high. The model successfully diagnosed 143 patients with the Glioma, only 7 of those were misdiagnosed as Meningioma. In Meningioma, classification was absolutely accurate, and all cases were correctly classified (146 cases). Likewise, there were no misclassification of the sample of 159 Normal cases with a perfect classification reading of 159 Normal. There were 147 correctly dealt with cases in Pituitary category with minor misclassifications, involving 2 Pituitary being predicted as Glioma and 1 Pituitary that were predicted as Meningioma. Altogether, the suggested model showed to be accurate, with few mistakes mainly among the diverse tumor types, reflecting its performance well in the detection of brain tumors. In addition, Figure 6 presents the ROC (above), the PR-AUC (below) curves of the proposed MobileDenseAttn model. Both the curves exhibit excellent behavior in the attributes of the four types of tumor: Glioma, Meningioma, Normal and Pituitary classification. All the ROC curves are quite near to the upper left corner, which means that the true positive and the false positive rate are high and low, respectively, and the AUC measures vary between 0.9993 (Glioma) and 1.0000 (Normal). These values of AUC imply the strong inter-class separation capacity of the model. On the same note, there is also high precision and recall on the PR-AUC curves too with the AUC values of the Glioma being 0.9980 and normal being 1.0000. This implies that the model is very accurate through different thresholds, and false positives and false negatives are minimal, which makes it especially suited when it comes to clinical applications.

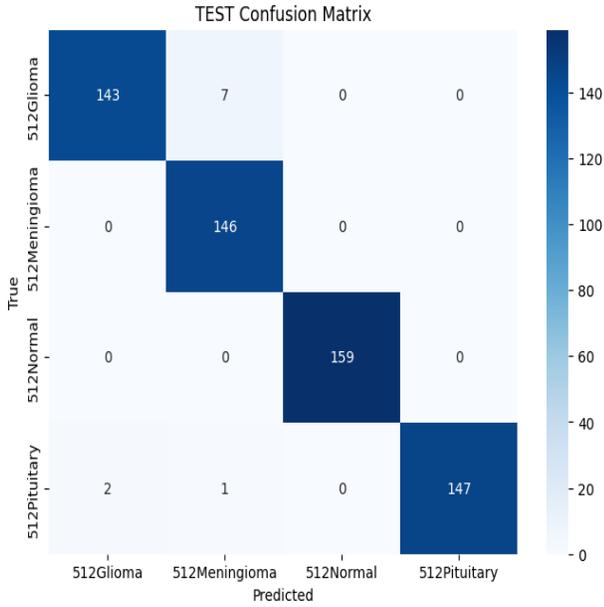

Fig. 5. Confusion Matrix of MobileDenseAttn.

## C. Model Reliability: Cohen's Kappa, Bootstrap Accuracy, and 95% Confidence Interval

The table III presents the test set performance of four models based on Cohen's Kappa, Bootstrap Accuracy, and the 95% Confidence Interval (CI). The Cohen Kappa is a metric to measure how well the predicted and actual labels match, and take chance agreement into account. VGG19 presents moderate scores with a Kappa of 0.9140, a Bootstrap accuracy of 93.55% and a CI, or parameters interval, of 91.60-95.51%. MobileNetV2 has a much stronger performance that Boostrapped accuracy is 97.69 percent and a Kappa of 0.9691 with CI 96.49 percent to 98.88 percent. DenseNet201 also records good results with a kappa of 0.9515, a Boostrap accuracy of 96.36 and a CI range between 94.87 and 97.86. Compared, the proposed MobileDenseAttn model yields a better result compared to the other models with a Kappa of 0.9840, and Bootstrap accuracy of 98.35%, and CI of 97.43% to 99.20%. These findings point to the proposed MobileDenseAttn as a model of the highest potential in regards to accuracy and generalization capacity, which demonstrates its appropriateness in real cases of use. The excellent results of the suggested model prove its competency to be used reliably in clinical settings, where one needs high reliability and generalization to make a good decision.

## D. Interpretability through Grad-CAM: Visualizing Tumor Detection in the Proposed MobileDenseAttn Model

In order to visualize how the proposed MobileDenseAttn model makes its decisions we used Grad-CAM (Gradient-weighted Class Activation Mapping) to individual backbones pre feature fusion, as shown in figure 7. Grad-CAM assists in making sense of what parts of the input picture each backbone is paying attention to in making predictions. The heatmaps generated in both backbones are almost similar to the actual region of the tumor, and therefore, the model is classifying and recognizing the significant tumor characteristics before fusion. Because the Grad-CAM cannot be explicitly used after the fusion of the features, we can assume that since both backbones target the affected areas, fused features will clearly help the proposed model to classify the images properly. This indicates that a combination of features of both networks contributes to better and dependable tumor detection.

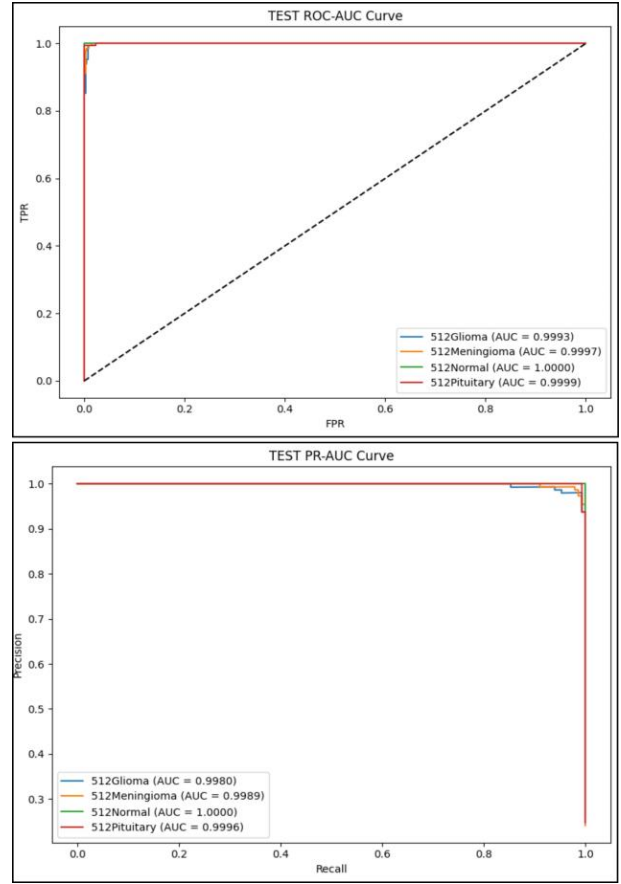

Fig. 6. ROC and PR AUC Curve of MobileDenseAttn.

TABLE III
TEST SET PERFORMANCE COMPARISON: COHEN'S KAPPA, BOOTSTRAP ACCURACY, AND 95% CI

| Model | Cohen's Kappa | Bootstrap Accuracy | 95% CI |
|---|---|---|---|
| VGG19 | 0.9140 | 93.55% | (0.9160, 0.9551) |
| MobileNetV2 | 0.9691 | 97.69% | (0.9649, 0.9888) |
| DenseNet201 | 0.9515 | 96.36% | (0.9487, 0.9786) |
| MobileDenseAttn | 0.9840 | 98.35% | (0.9743, 0.9920) |

## E. Comparative Analysis of Brain Tumor Detection Studies

Table IV shows our proposed MobileDenseAttn model is better than previous brain tumor detection techniques in that it exhibited the highest accuracy compared to the other reviewed studies. It has a dual-stream convolutional structure that allows robust multi-scale feature abstraction to yield great classifi-cation results. Furthermore, association with explainable AI methods include Grad-CAM which produces accurate and clinically meaningful heat maps to enable localization of tumors for better model interpretability. Notably, the model is also compact as regards inference time, proving it as fit in real-time clinical use.

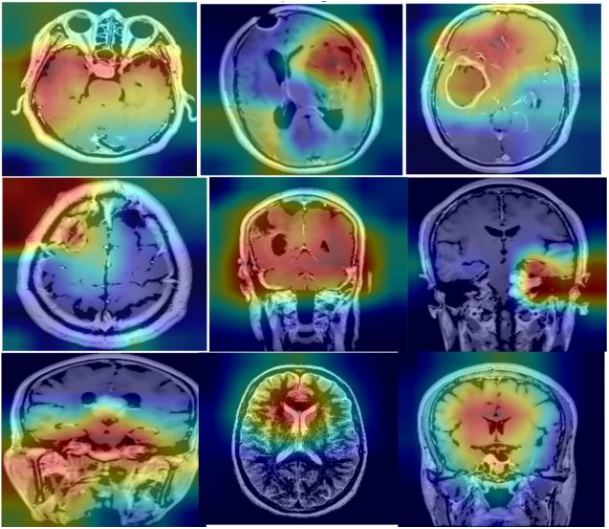

Fig. 7. GRAD Cam visualizaiton on both backbone of MobileDenseAttn.

TABLE IV
COMPARATIVE ANALYSIS OF BRAIN TUMOR DETECTION STUDIES

| Study | Model | Accuracy (%) |
|---|---|---|
| [10] | Hyperparameter-Optimized CNN | 97.0 |
| [11] | CNN with Machine Learning Integration | 92.86 |
| [13] | BrainNet (Transfer Learning CNN) | 97.71 |
| [15] | CNN (Compared with ResNet-50, VGG16, Inception) | 93.3 |
| Our Study | **MobileDenseAttn (Dual-stream CNN)** | **98.35** |

## IV. CONCLUSION

In short, MobileDenseAttn has introduced a potential solution to various fundamental challenges in brain tumor detection, which include accuracy, efficiency, and interpretability.MobileDenseAttn is an impressive model in terms of the test accuracy of 98.35 and a good value of F1-score of 0.9835 which uses a dual-stream framework that takes up the advantages of both MobileNetV2 and DenseNet201. The use of Grad-CAM would be helpful in gaining an understanding of how the model makes certain decisions and will create trust and transparency in clinical applications.

The high validity provided by 5-fold cross-validation shows that the model is stable and can be generalized to other datasets, which is why it may be applicable in the real-world environment. Additionally, the computational efficiency of MobileDenseAttn allows implementing the system in environments where resources are limited such that rapid and correct brain tumor is identified. On the whole, the study under review advances the medical imaging and artificial intelligence field, and it may have a beneficial impact on the current diagnostic results and patient care. Further work will be to train the model even more or use it on different imaging modalities and clinical situations, to make more of the technology in medical diagnosing.